\documentclass[10pt,twocolumn,letterpaper]{article}

\usepackage[pagenumbers]{cvpr}

\definecolor{cvprblue}{rgb}{0.21,0.49,0.74}
\usepackage[pagebackref,breaklinks,colorlinks,allcolors=cvprblue]{hyperref}
\usepackage{algorithm}
\usepackage[noend]{algpseudocode}
\usepackage{listings}
\usepackage{xcolor}
\title{Trainable Log-linear Sparse Attention for Efficient Diffusion Transformers}

\author{Yifan Zhou$^1$\ \hspace{12pt} Zeqi Xiao$^1$\ \hspace{12pt} Tianyi Wei$^1$\ \hspace{12pt} Shuai Yang $^{2}$\ \hspace{12pt} Xingang Pan$^{1}$\ \\
\normalsize{$^1 $S-Lab, Nanyang Technological University 
\hspace{12pt}
$^2$ Wangxuan Institute of Computer Technology, Peking University}\\
{\tt\small {\{yifan006, zeqi001, tianyi.wei, xingang.pan\}@ntu.edu.sg} \hspace{12pt} williamyang@pku.edu.cn}
}

\begin{document}

\twocolumn[{
    \renewcommand\twocolumn[1][]{#1}
    \maketitle
    \centering
    \includegraphics[width=\linewidth]{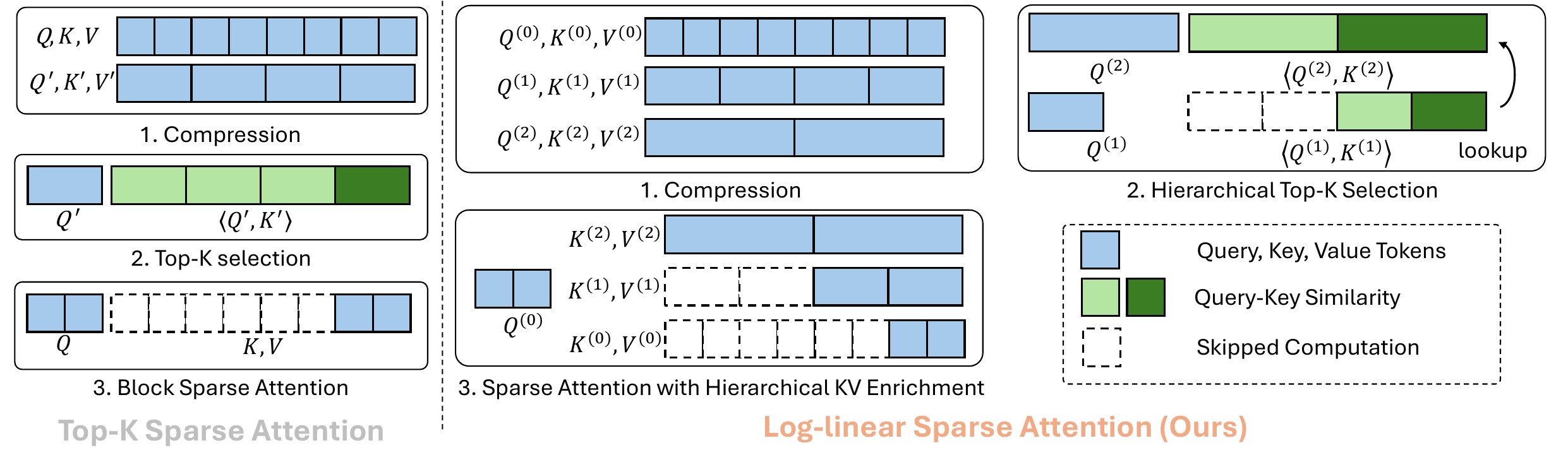}
    \captionof{figure}{Comparison between a general Top-$K$ sparse attention and our Log-linear Sparse Attention (LLSA). In the example, we use a token sequence of length $N=8$, block size $B=2$, Top-$K$ parameter $K=1$. To reduce the complexity of the selection stage from $O(N^2)$ to $O(N)$, we extend single-level selection to $O(\log N)$ levels. To achieve this, we compute the Top-$K$ of the full sequence on the coarsest level and recursively compute the sparse Top-$K$ on the remaining levels. To preserve the global context for attention, we enrich the key, value sets for each query with coarse tokens of length $O(K \log N)$ found in the selection stage.}
    \vspace{4mm}
    \label{fig:teaser}
}]

\maketitle
\begin{abstract}

Diffusion Transformers (DiTs) set the state of the art in visual generation, yet their quadratic self-attention cost fundamentally limits scaling to long token sequences. Recent Top-$K$ sparse attention approaches reduce the computation of DiTs by compressing tokens into block-wise representation and selecting a small set of relevant key blocks, but still suffer from (i) quadratic selection cost on compressed tokens and (ii) increasing $K$ required to maintain model quality as sequences grow. We identify that their inefficiency is due to the single-level design, as a single coarse level is insufficient to represent the global structure.
In this paper, we introduce \textbf{Log-linear Sparse Attention (LLSA)}, a trainable sparse attention mechanism for extremely long token sequences that reduces both selection and attention costs from quadratic to \textbf{log-linear complexity} by utilizing a hierarchical structure. LLSA performs hierarchical Top-$K$ selection, 
progressively adopting sparse Top-$K$ selection with the indices found at the previous level,
and introduces a Hierarchical KV Enrichment mechanism that preserves global context while using fewer tokens of different granularity during attention computation. 
To support efficient training, we develop a high-performance GPU implementation that uses only sparse indices for both the forward and backward passes, eliminating the need for dense attention masks.
We evaluate LLSA on high-resolution pixel-space image generation without using patchification and VAE encoding. LLSA accelerates attention inference by $ 28.27 \times$ and DiT training by $ 6.09 \times$ on $256 \times 256$ pixel token sequences, while maintaining generation quality. The results demonstrate that LLSA offers a promising direction for training long-sequence DiTs efficiently.

\end{abstract}    
\section{Introduction}
\label{sec:intro}

Diffusion Transformers (DiTs)~\cite{dit} have become the state-of-the-art backbone for visual generation tasks. As resolution and sequence length scale up, the dominant bottleneck lies in the quadratic complexity of full self-attention~\cite{transformer}, where the computation cost grows as $O(N^2)$ with token length $N$. In practice, this prevents DiTs from scaling to high-resolution images or long video sequences. For example, FLUX~\cite{flux} operates on a $64 \times 64$ latent image (4096 tokens), while Wan~2.1~\cite{wan} uses $21 \times 45 \times 80$ latent videos (75,600 tokens). Scaling to longer sequences requires a fundamentally more efficient attention mechanism.

Sparse attention has recently emerged as a promising alternative to full attention~\cite{zhangspargeattention, xi2025sparse, xia2025training}. A widely adopted variant is Top-$K$ block sparse attention~\cite{yuan2025native, lu2025moba, zhangspargeattention}, which operates in three stages: (1) compress query and key tokens into coarse representations that summarize the block-wise information; (2) compute the coarse similarity scores between compressed tokens, and select the Top-$K$ key blocks for each query block; (3) perform block sparse attention on the selected blocks. Although effective for moderate sequence lengths, this paradigm faces two major limitations when scaling further: (1) The selection stage still incurs quadratic cost on compressed tokens; (2) To maintain global context, prior methods set the sparsity as a constant and use a larger $K$ for longer sequences ~\cite{vsa, sla, vmoba}. These limitations arise from the \emph{single-level} design of existing Top-$K$ sparse attention: a single coarse-grained view is insufficient to represent global structure for long sequences.

A natural solution is to extend the single level into \emph{hierarchical} structure, where global information can be represented using only $O(\log N)$ coarse tokens of progressively coarser granularity.

Inspired by prior work showing that the dense attention matrix can be approximated by hierarchical coarse attention matrices ~\cite{zeng2022multi, zhu2021h}, we introduce \textbf{Log-linear Sparse Attention (LLSA)}, a trainable sparse attention mechanism that reduces attention complexity from quadratic to \emph{log-linear}. LLSA builds upon Top-$K$ sparse attention with two key innovations: (1) We compress query and key features across multiple logarithmic hierarchy levels and progressively perform Top-$K$ selection from coarse to fine. This hierarchical design reduces the complexity of the selection stage from $O(N^2)$ to $O(N)$. (2) Instead of using a large $K$ for longer sequences, we propose a \emph{Hierarchical KV Enrichment} mechanism that incorporates coarse key/value representations selected at higher hierarchy levels into attention computing. This preserves global context and mitigates information loss from sparsification, allowing LLSA to operate with significantly smaller $K$ and lower cost.

Efficient GPU implementation is essential for Top-$K$ sparse attention. In standard FlashAttention~\cite{flashattention1}, the sparse indices are represented as a binary mask. Constructing and processing this mask leads to quadratic memory and computation overhead. To avoid this, we implement a GPU-efficient Top-$K$ indexing algorithm that operates directly on sparse indices. In the forward pass, we gather only the selected key blocks. In the backward pass, we dynamically compute the reverse lookup of sparse indices via a lightweight sparse index transpose kernel. This ensures end-to-end log-linear complexity during training.

We evaluate LLSA on high-resolution pixel space image generation. Specifically, we train pixel DiTs without patchification and VAE encoding up to $256 \times 256$ (65,536) pixel tokens with one H200 GPU. LLSA significantly improves the training efficiency of full attention DiTs by $6.09 \times$ while maintaining generation quality. Compared to existing Top-$K$ sparse attention algorithms~\cite{sla, vsa}, LLSA achieves higher generation quality and training throughput. Thanks to Hierarchical KV Enrichment, LLSA maintains global context even with a significantly smaller $K$.
In our experiments, LLSA with $K=8$ outperforms prior Top-$K$ methods even when their $K$ is substantially larger ($K=20$ or $K=32$), highlighting the practical efficiency of our design. We further integrate LLSA into PixelFlow~\cite{chen2025pixelflow} to show its capacity on ImageNet~\cite{imagenet}. Moreover, our backward kernel achieves nearly constant throughput across different sequence lengths, confirming the linear complexity of our implementation.

In summary, the contributions are listed as follows.
\begin{itemize}
    \item We propose Log-linear Sparse Attention (LLSA), a trainable attention mechanism that scales DiTs to long sequences with log-linear complexity and comparable quality. 
    \item We develop a high-performance GPU implementation for Top-$K$ indexing that eliminates dense mask construction in both forward and backward passes.
    \item Extensive empirical results showing that LLSA consistently outperforms prior Top-$K$ sparse attention methods, in both quality and efficiency, especially at small $K$, and scales effectively to high-resolution pixel DiTs and large-scale datasets.
\end{itemize}
\section{Related Works}
\label{sec:related_works}

\subsection{Sparse Attention in DiTs}
Early attempts mainly adopt training-free strategies that prune irrelevant tokens using predefined sparsity patterns~\cite{chen2025sparse, xi2025sparse, yang2025sparse} or dynamic search rules~\cite{zhangspargeattention, jiang2024minference, xia2025training}. More recently, inspired by trainable sparse attention in NLP~\cite{lu2025moba, yuan2025native}, several methods, such as  VMoBA~\cite{vmoba}, VSA~\cite{vsa}, and SLA~\cite{sla}, introduce learnable sparse attention for efficient DiT training. These approaches rely on a single-level block selection mechanism that identifies important token blocks at one resolution. Our work extends this idea to a multi-level hierarchical selection scheme, enabling substantially better scalability for long-sequence attention.

\subsection{Log-linear Attention}
Log-linear attention approaches can be categorized by their strategies for determining the relevant keys for each query. Several methods use static rules based on absolute positions. H-Transformer~\cite{zhu2021h} constructs a fixed hierarchical decomposition of the attention matrix; Fast Multipole Attention~\cite{kang2023fast} applies Fast Multipole Method to use different resolutions according to the distance between the query and key tokens. Radial Attention~\cite{li2025radial} imposes a static attention mask according to the pattern of the video data. Log-linear Attention~\cite{guo2025log} derives a linear-attention variant that maintains logarithmic hidden states using a Fenwick tree. Other methods rely on dynamic key selection. Reformer~\cite{kitaev2020reformer} efficiently clusters the queries and keys using locality-sensitive hashing. The most relevant prior work is Multi-resolution Attention~\cite{zeng2022multi}, which performs hierarchical Top-$K$ selection. However, its contribution is mainly theoretical and does not provide a high-performance GPU implementation for extremely long token sequences. In contrast, our method's implementation is based on block sparse attention~\cite{flashattention1}, and is validated on long-sequence DiT training.

\subsection{Pixel-space DiTs}

Due to computational constraints, the original DiT~\cite{dit} operates in latent space~\cite{ldm}, relying on VAE encoders and patchification to reduce token length. Pixel-space DiTs therefore require either more efficient diffusion processes or architectural modifications involving aggressive downsampling.  PixelFlow~\cite{chen2025pixelflow} splits the diffusion process into multiple stages of different resolutions and uses a universal DiT for all stages. PixNerd~\cite{pixnerd} integrates a coordinate-based MLP to decode pixel-level output of a large-patch DiT. HDiT~\cite{hdit} introduces a U-shaped DiT with progressive downsampling and upsampling. To our knowledge, no prior work has trained a pure pixel-space DiT achieving state-of-the-art performance on high-resolution data without any input downsampling. In this paper, we verify our LLSA  on a lightweight pixel-space DiT trained on a high-resolution dataset, demonstrating that our attention mechanism matches the quality of full attention while significantly improving efficiency.

\section{Background}
\label{sec:background}

In this section, we introduce a simplified implementation of Block Sparse FlashAttention~\cite{flashattention1, flashattention2} without the scaling factor and safe softmax. We then analyze the computing complexity of Block Sparse FlashAttention with Top-$K$ selection.

\subsection{Block Sparse FlashAttention}

Given query, key, value features $\mathbf{Q}, \mathbf{K}, \mathbf{V} \in \mathbb{R}^{N\times d}$, where $N$ is the length of the token sequence and $d$ is the feature dimension, the attention output $\mathbf{O} \in \mathbb{R}^{N\times d}$ is:
\begin{equation}
    \mathbf{P} = \text{softmax}(\mathbf{Q} \mathbf{K}^{\intercal}), \mathbf{O}=\mathbf{P}\mathbf{V}.
\end{equation}

To avoid storing the large intermediate matrix $\mathbf{P} \in \mathbb{R}^{N\times N}$ in high bandwidth memory (HBM), FlashAttention uses tiling to compute output block-by-block. Specifically,  $\mathbf{Q}, \mathbf{K}, \mathbf{V}, \mathbf{O}$ are divided into $T$ blocks of block size $B$ such that $T=\lceil \frac{N}{B} \rceil$, \eg, $\mathbf{Q} = [\mathbf{Q}_1, ..., \mathbf{Q}_T], \mathbf{Q}_i \in \mathbb{R}^{B\times d}$. For each query block $\mathbf{Q}_i$, we compute $\mathbf{O}_{i}$ as:
%
\begin{align}
    \mathbf{P}_{i, j} &= \exp(\mathbf{Q_i} \mathbf{K_j}^{\intercal}) \in \mathbb{R}^{B\times B} \\
    \tilde{\mathbf{O}_i} &= \sum_{j=1}^{T}{\mathbf{P}_{i, j}\mathbf{V}_j} \in \mathbb{R}^{B\times d}\\
    l_i &= \sum_{j=1}^{T}{\text{rowsum}{(\mathbf{P}_{i, j}})} \in \mathbb{R}^{B} \\
    \mathbf{O}_{i} &= \tilde{\mathbf{O}_i} \oslash l_i,
\end{align}
where $\oslash$ denotes element-wise division with $l_i$ broadcast across columns. In implementation, we compute $\mathbf{P}_j$ in the $j$-th iteration and accumulate $\tilde{\mathbf{O}_i}$ and $l_i$ across the loop. 

FlashAttention can be natively extended to a sparse version if we know that the attention between the $i$-th $\mathbf{Q}$ block and the $j$-th $\mathbf{K}$ block can be skipped: Given a binary sparsity mask $\mathbf{M} \in \mathbb{R}^{T\times T}$, the iteration that computes $\mathbf{P}_{i, j}$ is skipped if $\mathbf{M}_{i, j} = 0$. The main challenge lies in determining $\mathbf{M}$. A common method for identifying the $\mathbf{M}$ is the Top-$K$ selection approach.

\subsection{Complexity of Top-$K$ Sparse Attention}

The Top-$K$ selection consists of three steps. First, we compress the inputs in each block via mean pooling and obtain $\mathbf{Q}',\mathbf{K}' \in \mathbb{R}^{T \times d}$, where $\mathbf{Q}'_{i},\mathbf{K}'_{j}$ is the summarization of $\mathbf{Q}_{i}, \mathbf{K}_{j}$. Then, we compute the full attention score $S = \mathbf{Q}' {\mathbf{K}'}^{\intercal}$ on the compressed representations. Finally, we sort the attention score for each query block and mark the Top-$K$ key blocks as valid.

With Top-$K$ selection, the Top-$K$ sparse attention can be divided into two stages: a selection stage that computes similarities on compressed tokens and performs Top-$K$ to obtain $\mathbf{M}$, and a sparse attention stage that operates on the selected blocks. The selection stage computes pairwise scores among the $T$ coarse tokens and applies Top-$K$ selection on the scores, resulting in a complexity of $O(T^2 d) + O(T^2K)= O(N^2 B^{-2} (d+K))$. In contrast, the sparse attention stage processes only $K$ blocks per query, with a complexity of $O(NKBd)$. 

The total cost is thus
    $O(N^2B^{-2}(d+K)) + O(KNBd)$,
However, as we scale to longer sequences, the quadratic term in the selection stage, $O(N^2)$, dominates the total running time. This means that even though the sparse attention computation grows only linearly with $N$, the overall cost is still bottlenecked by the $O(N^2)$ selection stage. As a result, existing Top-$K$ sparse attention methods fail to maintain efficiency when $N$ becomes large.

\section{Log-linear Sparse Attention Mechanism}
\label{sec:method}

\setlength{\textfloatsep}{8pt}
\begin{algorithm}[t]
  \caption{Log-linear Sparse Attention}
  \textbf{Input:} Features $\mathbf{Q}, \mathbf{K}, \mathbf{V} \in \mathbb{R}^{N\times d}$, block size $B$, Top-$K$ parameter $K$, number of coarse levels $L$, number of KV Enrichment levels $L_e$. \\
  \textbf{Output:} Attention output $\mathbf{O}$.
  \begin{algorithmic}[1]
    \State $\mathbf{Q}^{(0)}, \mathbf{K}^{(0)}, \mathbf{V}^{(0)} = \mathbf{Q}, \mathbf{K}, \mathbf{V}$

    \For{$l = 1$ \textbf{to} $L$} \Comment{Hierarchical Compression}
        \State $\mathbf{Q}^{(l)},\mathbf{K}^{(l)},\mathbf{V}^{(l)} = \text{pool}_B(\mathbf{Q}^{(l-1)}, \mathbf{K}^{(l-1)}, \mathbf{V}^{(l-1)})$
    \EndFor
    \State \textbf{end for} 

    \State $I^{(L)} = [1, ..., \frac{N}{B^{L}}] $
    

    \For{$l = L$ \textbf{to} $1$} \Comment{Hierarchical Top-$K$ Selection}
        \State Divide $\mathbf{Q}^{(l)}, \mathbf{K}^{(l)}, \mathbf{V}^{(l)}$ into $T^{(l)}= \frac{N}{B^{l+1}}$ blocks $\{\mathbf{Q}^{(l)}_t\}, \{\mathbf{K}^{(l)}_t\}, \{\mathbf{V}^{(l)}_t\}$

        \For{$i = 1$ \textbf{to} $T$}
            \State Gather $\mathbf{K}^{(l)}_i$ from $\{\mathbf{K}^{(l)}_t\}$ with indices $I^{(l)}_i$
            \State $\mathbf{S}^{(l)}_i = \mathbf{Q}^{(l)}_i{\mathbf{K}^{(l)}_i}^{\intercal}$
            \State $I^{(l-1)}_i = \text{topk\_indices}(\mathbf{S}^{(l)}_i,K)$

        \EndFor 
        \State \textbf{end for} 
        
    \EndFor
    \State \textbf{end for}

    \State Divide $\mathbf{Q}^{(0)}, \mathbf{K}^{(0)}, \mathbf{V}^{(0)}$ into $T= \frac{N}{B}$ blocks $\{\mathbf{Q}^{(0)}_t\}, \{\mathbf{K}^{(0)}_t\}, \{\mathbf{V}^{(0)}_t\}$

    \For{$i = 1$ \textbf{to} $T$} \Comment{Attention}
        
        \State $\mathbf{K}_c \gets \{\}, \mathbf{V}_c \gets \{\}$ 
        \For{$l = 0$ \textbf{to} $L_e$} \Comment{Hierarchical KV Enrichment}
            \State Gather $\mathbf{K}^{(l)}_i, \mathbf{V}^{(l)}_i$ from $\{\mathbf{K}^{(l)}_t\}, \{\mathbf{V}^{(l)}_t\}$ with indices $I^{(l)}_i$
            \State $\mathbf{W}^{(l)}=B^l$\Comment{KV Reweighting}
            \State Append $\mathbf{K}^{(l)}_i \cdot \mathbf{W}^{(l)}, \mathbf{V}^{(l)}_i \cdot \mathbf{W}^{(l)}$ to $\mathbf{K}_c, \mathbf{V}_c$
            
        \EndFor 
        \State \textbf{end for}

        \State $\mathbf{O}_{i} = \text{FlashAttention}(\mathbf{Q}^{(0)}_i, \mathbf{K}_c, \mathbf{V}_c)$

    \EndFor 
    \State \textbf{end for} 

    \State \textbf{return} $\mathbf{O} = \{\mathbf{O}_{i}\}$

  \end{algorithmic}
  \label{alg:algorithm1}
\end{algorithm}

LLSA aims to reduce the complexity of Top-$K$ sparse attention from $O(N^2)$ to $O(N \log N)$. This requires both an efficient and effective algorithm, as well as a high-performance GPU kernel implementation for sparse operations. We introduce the algorithm of LLSA in Sec.~\ref{sec:method1}, the complexity analysis in Sec.~~\ref{sec:method2}, and the implementation of Top-$K$ kernels in Sec.~\ref{sec:method3}  

\subsection{Log-linear Sparse Attention}
\label{sec:method1}

As discussed in Sec.~\ref{sec:background}, a general Top-$K$ sparse attention can be divided into three steps: compression, Top-$K$ selection, and sparse attention. To reduce the complexity of the attention, we extend the single-level compression into logarithmic hierarchy compression and apply sparse Top-$K$ selection in a coarse-to-fine manner. Since the receptive field of each query is reduced by the sparse selection, we propose Hierarchical KV Enrichment to preserve the global context. To ensure the coarse tokens have correct importance, we apply KV Reweighting to the coarse tokens during attention computation. The full algorithm is shown in Alg.~\ref{alg:algorithm1}. For simplicity, we assume the shapes of $\mathbf{Q}, \mathbf{K}, \mathbf{V}$ are the same, the token length $N$ is divisible by the largest block size $B^L$, and the parameter $K$ is the same for all coarse levels. 

\noindent \textbf{Hierarchical Compression.}
Given attention inputs $\mathbf{Q}, \mathbf{K}, \mathbf{V} \in \mathbb{R}^{N\times d}$, block size $B$, number of levels $L=\lfloor \log_BN - 1\rfloor$ (as we need to reserve enough tokens for tiling on the coarsest level), we compute hierarchical representation $\mathbf{Q}^{(l)}, \mathbf{K}^{(l)}, \mathbf{V}^{(l)} \in \mathbb{R}^{N/B^l\times d}$. We set $\mathbf{Q}, \mathbf{K}, \mathbf{V}$ as the finest tokens $\mathbf{Q}^{(0)}, \mathbf{K}^{(0)}, \mathbf{V}^{(0)}$ and recursively downsample the features via mean pooling. As a result, a coarse token in $l$-th level is a summarization of $B$ tokens in $(l-1)$-th level.

\noindent \textbf{Hierarchical Top-$K$ Selection.}
In this step, we aim to obtain the hierarchical Top-$K$ indices $I^{(l)}$ for all levels. We introduce a hyperparameter $K$ that denotes the number of activated $\mathbf{K}, \mathbf{V}$ tokens in each level. Similar to block sparse FlashAttention~\cite{flashattention1}, we assume neighboring $B$ $\mathbf{Q}$ tokens share the same sparse pattern. Therefore, for level $l$, the shape of sparse indices $I^{(l)}$ is $N/B^{l+1} \times K$. 

In hierarchical Top-$K$ selection, we first compute the full similarity matrix $\mathbf{S}^{(L)}$ of coarsest tokens $\mathbf{Q}^{(L)}, \mathbf{K}^{(L)}$ by $\mathbf{S}^{(L)} = \mathbf{Q}^{(L)}{\mathbf{K}^{(L)}}^{\intercal}$. We then compute the Top-$K$ indices of $\mathbf{S}^{(L)}$ for each query token. The indices indicate that one query block at the $L-1$ level has $K B$ key candidates. Starting from the second coarsest level, we gather the keys from the sparse indices and compute the similarity and sparse Top-$K$ indices from $K B$ candidates for the next level recursively.
The total complexity of this stage is $O(N)$, and the proof will be given later.

\noindent \textbf{Hierarchical KV Enrichment.}
In the final step, we aim to compute the final attention output using the hierarchical Top-$K$ indices obtained in the previous step. In addition to the finest sparse tokens $\mathbf{K}^{(0)}, \mathbf{V}^{(0)}$, we append the sparse tokens $\mathbf{K}^{(l)}_i, \mathbf{V}^{(l)}_i$ from all levels selected in the previous step to the final key, value sets for each query block. The coarse key, value tokens contain the information of the global context, filling up the information loss caused by the sparse selection. Since the activated tokens are obtained from sparse indices and the number of levels is $O(\log N)$, the number of activated tokens is $O(N\log N)$. 

In practice, as shown in Alg.~\ref{alg:algorithm1}, we introduce a hyperparameter $L_e$ to determine the number of KV enrichment levels. For example, when $L_e = 0$, then we do not append the coarse token. When $L_e = 1$, we append the coarse tokens from the second finest level. We set $L_e=L$ by default.

\noindent \textbf{KV Reweighting.}
The design of the Hierarchical KV Enrichment mechanism introduced in the last section is not optimal: the coarse tokens contain more information, but their importance is equivalent to that of the finest tokens, which is not sufficient. To address this, we assume the finest tokens corresponding to one coarse token can be roughly recovered by nearest upsample. Therefore, the importance of each token should be proportional to its block size. \eg, a coarse token obtained by averaging $16$ tokens should have a weight of $16$. In practice, we add a weighting term $\mathbf{W}^{(l)}$ to $\mathbf{K}^{(l)}, \mathbf{V}^{(l)}$ such that $\mathbf{W}^{(l)}=B^l$. This design enhances the model quality without increasing the training overhead.

\subsection{Complexity Analysis}
\label{sec:method2}
LLSA consists of two main components: hierarchical Top-$K$ selection and sparse attention with Hierarchical KV Enrichment.

\noindent \textbf{Selection Stage.}
At the hierarchy level $l$, let $N^{(l)} = N / B^l$ denote the sequence length. For each level, we compute similarity scores only between the $N^{(l)} / B$ fine query blocks and the $KB$ candidate key blocks identified from the coarser level. The total cost across $L - 1$ levels (except the coarsest level) is therefore
\begin{equation}
    \sum_{l=0}^{L-1} O \left(\frac{N}{B^{l+1}} KB\right)
    = O\!\left(
    NK \sum_{l=0}^{L-1} B^{-l}
    \right)
    = O(NK),
\end{equation}
because the geometric series $\sum_{l=1}^{L} B^{-l} < \frac{B}{B-1}$ converges to a constant independent of $N$ and $K$. Thus, the selection stage runs in $O(NK)$ instead of the $O(N^2)$ cost of the previous Top-$K$ sparse attention.\footnote{We only analyze the complexity of similarity matrix computation for simplicity. The complexity of Top-$K$ depends on the algorithm used in implementation. If both $B$ and $K$ are constant, then the complexity of Top-$K$ is still $O(N)$.}

\noindent \textbf{Sparse Attention Stage.}
For each query, LLSA attends to (i) the $K$ fine-level Top-$K$ keys and (ii) a logarithmic number of enriched coarse KV tokens collected from all hierarchy levels. The number of enriched KV tokens is therefore $O(K \log N)$. Computing attention over these candidates gives a complexity of $O(N K \log N)$.

\noindent \textbf{Overall Complexity.}
Combining the two components, the total complexity of LLSA is $O(NK) \;+\; O(NK \log N) \;=\; O(NK \log N)$.
Since $K$ is kept constant regardless of $N$, the overall complexity becomes $O(N \log N) $,
which is asymptotically more efficient than the $O(N^2)$ cost of full attention and the $O(N^2)$ dominant term in previous Top-$K$ sparse attention methods.

\subsection{Efficient Kernel Implementation}
\label{sec:method3}

\setlength{\textfloatsep}{8pt}
\begin{algorithm}[t]
  \caption{Sparse Top-$K$ Indices Transposition}
  \textbf{Input:} Indices $\mathbf{I} \in \mathbb{R}^{T\times K}$,  Top-$K$ parameter $K$. \\
  \textbf{Output:} Flat sparse query indices $\mathbf{I}_q \in \mathbb{R}^{T \cdot K}$, cumulative key offsets $\mathbf{C} \in \mathbb{R}^{T+1}$.
  \begin{algorithmic}[1]

    \State $\mathbf{C'} \gets \mathbf{0} \in \mathbb{R}^{N+1}$
    \State $\mathbf{I}_q \gets \mathbf{0} \in \mathbb{R}^{T \cdot K}$

    \For{$i = 1$ \textbf{to} $T$} \Comment{Count}
        
        \For{$j = 1$ \textbf{to} $K$} 
            \State $I \gets \mathbf{I}_{i, j} + 1$
            \State $\mathbf{C}_{I}'\gets \mathbf{C}_{I}' + 1$
            
        \EndFor 
        \State \textbf{end for}

    \EndFor 
    \State \textbf{end for} 

    \State $\mathbf{C}$ $\gets$ \text{cumsum}($\mathbf{C'}$)

    \State $\mathbf{C}_O \gets \mathbf{0} \in \mathbb{R}^{N}$ \Comment{Temporary offsets}

    \For{$i = 1$ \textbf{to} $T$} \Comment{Indices transpose}
        \For{$j = 1$ \textbf{to} $K$} 
            \State $I \gets \mathbf{I}_{i, j}$
            \State $(\mathbf{C}_O)_{I} \gets (\mathbf{C}_O)_{I} + 1$
            \State $I \gets \mathbf{C}_{I}+(\mathbf{C}_O)_{I}$ 
            \State ${(\mathbf{I}_q)}_{I} \gets i$
            
        \EndFor 
        \State \textbf{end for}
        
    \EndFor 
    \State \textbf{end for} 

    \State \textbf{return} $\mathbf{I}_q, \mathbf{C}$

  \end{algorithmic}
  \label{alg:algorithm2}
\end{algorithm}

To ensure the algorithms align with theoretical computational complexity, we must implement a high-performance, parallel program adapted to modern GPUs. For most query-major operations (\eg, attention forward, query gradient backward), we only need to replace the full key, value traversal with sparse key, value gathering in FlashAttention~\cite{flashattention2} forward loop. However, the implementation of key, value backward, a key-major operation, is not trivial since we do not have the sparse query block indices for each key block. Existing Top-$K$ sparse attention methods \cite{vsa, sla} maintain a sparse mask of shape $T \times T$ for backward. However, this implementation increases the complexity of the algorithm from $O(T)$ to $O(T^2)$. 

To avoid constructing the sparse mask and achieving optimal complexity, we implement an efficient sparse Top-$K$ index transposition algorithm following the classic CSR-to-CSC scan-based sparse matrix transpose algorithm~\cite{gustavson1978two} and its parallel variant~\cite{merrill2016merge}. The algorithm for one level is illustrated in Alg.~\ref{alg:algorithm2}. The hierarchical Top-$K$ transposition can reuse the same program for each level.  

The overall idea of the algorithm is that we save all the query indices for each key of variable length into a flat vector $\mathbf{I}_q \in \mathbb{R}^{T K}$, and we use the cumulative key offsets $\mathbf{C} \in \mathbb{R}^{T+1}$ to obtain the start and end index for each key in $\mathbf{I}_q$. 

To obtain $\mathbf{I}_q$ and $\mathbf{C}$, we first count the number of relevant queries for each key into $\mathbf{C'}$, and the prefix-sum of $\mathbf{C'}$ is the cumulative key offsets $\mathbf{C}$. Given $\mathbf{C}$, we traverse the top-$K$ indices again and write the reverse mapping into $\mathbf{I}_q$ with the help of a cumulative offset vector $\mathbf{C}_O$. Note that the update of both $\mathbf{C'}$ and $\mathbf{C}_O$ requires atomic additions in parallel programming. Because the number of selected keys $K$ is small, the atomic additions are sparsely distributed across memory locations. Consequently, the probability of inter-program write conflicts is extremely low, and the overhead induced by atomic contention becomes negligible.

The key-major data structure $\mathbf{I}_q, \mathbf{C}$ allow us to perform the key, value gradient backward with sparse–dense matrix multiplication (SpMM) accumulation, \ie, we gather the sparse query indices $(\mathbf{I}_q)_{[C_{i}: C_{i+1})}$ for the $i$-th key and perform standard backward algorithm for key and value.

\section{Experiments}
\label{sec:experiments}

\subsection{Pixel DiT Implementation}
\label{sec:exp_imple}

To validate that LLSA effectively reduces computational complexity, we train Pixel DiTs on the task of generating long pixel sequences \emph{without} patchification or VAE encoding. To adapt LLSA to 2D visual data, we introduce an \textbf{index-reordering} scheme, and to accelerate the convergence of DiT on long token sequences, we apply a modified noise scheduler with \textbf{noise rescaling} and enable \textbf{low-resolution pretraining}.

\begin{figure}[t]
\centering
\includegraphics[width=\linewidth]{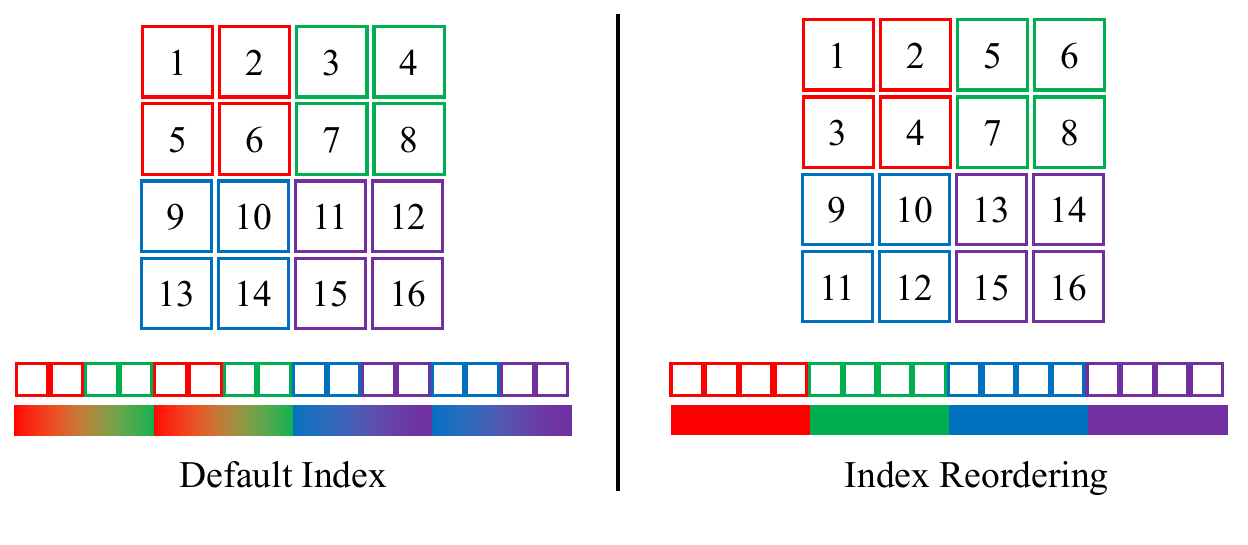}\vspace{-2mm}
\caption{Illustration of index reordering. The default raster indices do not effectively cluster similar pixels during 1D pooling, while using index ordering guarantees that similar pixels receive neighboring 1D indices.}

\label{fig:reordering}
\end{figure}

\noindent \textbf{Index Reordering.}
To extend LLSA beyond 1D sequence modalities, the only modification required is to reorder tokens so that spatially adjacent pixels become neighbors in the flattened 1D sequence. Following prior sparse-attention approaches~\cite{zhang2025training, zhangspargeattention}, we apply index reordering once at the beginning of the DiT. Specifically, pixels within each patch of size $2^i$ are grouped as consecutive tokens in the flattened sequence, as illustrated in Fig.~\ref{fig:reordering}.  
This preserves local continuity in 2D layouts while maintaining compatibility with the 1D hierarchical Top-$K$ routing in LLSA.

\noindent \textbf{Noise Rescaling.}
Diffusion models require stronger noise perturbations for higher-resolution images to maintain a consistent signal-to-noise ratio (SNR)~\cite{sd3, chen2023importance, hoogeboom2023simple}.  
Inspired by the input-scaling strategy of~\cite{chen2023importance}, we accelerate pixel-space DiT training by instead scaling the noise. We introduce a rescale factor $s$ into the flow matching scheduler~\cite{flow_match_1, flow_match_2},

\begin{equation}
    \mathbf{x}_t = (1-t)\mathbf{x}_0+s \cdot t\epsilon,
\end{equation}
where $\mathbf{x}_t \in \mathbb{R}^{n\times n}$. We set $s = n / 64$ for images larger than $64\times64$, aligning their effective SNR with that of $64\times64$ images and thus stabilizing training at higher resolutions.

\noindent \textbf{Pretraining.} Consistent with large text-to-image DiTs~\cite{sd3}, we initialize high-resolution models from low-resolution pretrained weights. This design further reduces the training time. All detailed settings, including data resolution schedules and optimization parameters, are provided in the Appendix.

\noindent \textbf{Metrics.}
For the quality metric, we evaluate the FID~\cite{fid} using 10,000 samples generated by 20 diffusion steps. For model efficiency, we report the training throughput measured as $10^3$ pixel tokens per second on a single H200 GPU.

\subsection{Ablation Study}
\label{sec:abl}

\noindent \textbf{Training Settings.}
In the ablation study, we use a pixel DiT-S~\cite{dit} with flow matching object~\cite{sit} on $128 \times 128$ FFHQ~\cite{stylegan} dataset. We incorporate RoPE~\cite{rope} and QK-Norm~\cite{qknorm} to align the standard DiT with other advanced DiT architectures~\cite{chen2025pixelflow, hdit}. In each experiment, we train the model for 20 epochs with the same training configurations. If not specified, we set $K=8$ and $B=16$ by default for all Top-$K$ sparse attention methods. For a token length of $N$, the maximum level should be $\lfloor \log_B N - 1\rfloor$. 

\begin{table} [t]
\caption{Ablation study results of Log-linear Sparse Attention}\vspace{-2mm}
\label{tab:abl}

\centering
\subfloat[Attention Type]{%
  \begin{minipage}[t]{0.8\linewidth}
\centering
\resizebox{\linewidth}{!}{
\begin{tabular}{l|c|c}
\toprule
Configuration & FID & Throughput  \\
\midrule 
Full Attention & 24.91 & 188.88 \\
\midrule 
Top-$K$ Attention ($L=1$) & 28.21 & 483.91 \\
+ KV Enrichment ($L_e=1$) & 26.09 & 302.92 \\
+ KV Reweighting & 24.18 & 302.92 \\
\midrule 
Top-$K$ Attention ($L=2$)  & 27.98 & 500.38 \\
+ KV Enrichment ($L_e=2$) & 25.31 & 436.40 \\
+ KV Reweighting & 24.37 & 436.40 \\
\bottomrule
\end{tabular}
}
\end{minipage}
\label{tab:abl_1}
}
\par\vspace{0.5em}

\subfloat[Block Size]{%
  \begin{minipage}[t]{0.40\linewidth}
\centering
\resizebox{1.1\linewidth}{!}{
\begin{tabular}{c|c|c|c}
\toprule
$B$ & $K$ & FID & Throughput  \\
\midrule 
16 & 8 & 28.21 & 483.91 \\
64 & 2 & 31.33 & 584.54 \\
\midrule 
16 & 32 & 25.88 & 357.95 \\
64 & 8 & 26.52 & 547.69 \\
\bottomrule
\end{tabular}
}
\end{minipage}
\label{tab:abl_2}
}
\hfill
\subfloat[Top-$K$]{%
  \begin{minipage}[t]{0.52\linewidth}
\centering
\resizebox{1.0\linewidth}{!}{
\begin{tabular}{l|c|c|c}
\toprule
Config & $K$ & FID & Throughput \\
\midrule 
LLSA & 8 & 24.37 & 436.40 \\
\midrule 
Baseline & 8 & 28.21 & 483.91 \\
 & 16 & 27.23 & 436.40 \\
& 32 & 25.88 & 357.95 \\
\bottomrule
\end{tabular}
}
\end{minipage}
\label{tab:abl_3}
}

\end{table}

\noindent \textbf{Attention Designs.}
We verify our attention designs in Table~\ref{tab:abl_1}. We test the effectiveness of KV Enrichment and KV Reweighting with coarse level $L=1$ and $L=2$. Solely with Top-$K$ attention can hardly match the performance of full attention. Enabling KV Enrichment improves FID, and correctly setting the importance for coarse tokens further enhances the model quality, making it even better than the full attention baseline. However, the single-level attention with KV Enrichment still has an $O((N/B)^2)$ complexity in both the selection and attention stages. Extending to a multi-level version increases the throughput significantly with a minor quality drop.

\noindent \textbf{Lower Block Size is Better.} We conduct two groups of experiments with Top-$K$ attention $(L=1)$ in Table~\ref{tab:abl_2} to explore the influence of block size. In each group, we test $B=16$ and $B=64$ settings with the same relevant token numbers. A larger block size increases the throughput by a considerable amount because it reduces the overhead of the selection stage, but its quality is much worse than that of the small block size variant. Since the complexity of selection is reduced by our hierarchical design, unlike prior works that use a larger block size~\cite{zhangspargeattention, vmoba, sla, vsa}, we use $B=16$ as the default setting.

\noindent \textbf{Small $K$ is Enough.}
We compare the final LLSA $(L=L_e=2)$ to the baseline sparse attention ($L=1$) of varying $K$. The baseline requires a higher $K$ to match the quality of the full attention. In contrast, LLSA uses only $K=8$ and outperforms the $K=32$ baseline in both quality and efficiency. This indicates that the Hierarchical KV Enchirment ensures the LLSA to preserve global context with a relatively small $K$. 

\subsection{Image Generation Benchmark}
\label{sec:comparison}

\noindent \textbf{Comparison with Baselines.} We compare LLSA with two trainable Top-$K$ sparse attention methods, VSA~\cite{vsa} and SLA~\cite{sla}, on FFHQ \(128\times128\) and \(256\times256\) image generation using a DiT-S backbone.
Both methods rely on a single-level block selection scheme but process the coarse tokens differently:  
VSA performs full attention on the compressed (coarse) $\mathbf{Q}, \mathbf{K}, \mathbf{V}$ tokens and adds the result to the sparse attention output, while SLA introduces an additional linear-attention branch to handle unselected tokens.

To ensure a fair comparison, we adjust the Top-$K$ values so that \emph{all methods attend to approximately the same number of key blocks} during sparse attention.  
Because LLSA incorporates multi-level coarse-to-fine KV enrichment, a query with $K=8$ effectively attends to more blocks (\eg, for \(128\times128\) images, \(8 + 8 + 128^2 / 16^3 = 20\) blocks).  
Thus, to provide VSA and SLA with equal capacity in the sparse attention stage, we set \(K=20\) for the $128\times128$ image and \(K=32 \) for the $256\times256$ image, for both methods.  
This increases the strength of the baselines, making our evaluation conservative: LLSA must match or exceed the performance of baselines that operate with a larger \(K\).

Table~\ref{tab:comparison} summarizes the results.  
Across both resolutions, LLSA achieves the best FID and the highest training throughput.  
This validates the effectiveness of hierarchical KV enrichment, which preserves the single-branch structure of standard attention. In contrast, the baselines rely on multiple attention branches (coarse attention or linear attention), which may distort the original attention formulation.

\begin{table} []

\caption{Comparison of LLSA with other trainable Top-$K$ sparse attention. We show the FID and training throughput for FFHQ-128 (20 epochs) and FFHQ-256 (10 epochs). Training throughput is measured as $10^3$ pixel tokens per second on a single H200 GPU.}\vspace{-2mm}
\label{tab:comparison}
\centering
\resizebox{0.9\linewidth}{!}{
\centering

\begin{tabular}{l|c|c|c|c}
\toprule
 & \multicolumn{2}{|c|}{$128\times128$} & \multicolumn{2}{|c}{$256\times256$} \\
 \midrule 
Method & FID & Throughput  & FID & Throughput \\
\midrule 
Full Attention & 24.91 & 188.88 & 38.77 & 61.64\\
\midrule 
VSA & 26.91 & 421.02 & 40.69 & 341.94 \\
SLA & 25.73 & 365.48 & 39.98 & 304.85 \\
LLSA & \textbf{24.37} & \textbf{436.40} & \textbf{39.29} & \textbf{375.34} \\

\bottomrule
\end{tabular}}\vspace{-0mm}
\end{table}

\begin{table} []
\caption{PixelFlow ImageNet-256 benchmark on different sparse attention methods trained for 10 epochs. FID and Inception Score are computed on 10,000 samples with PixelFlow's official script. Training throughput is measured as images per second on a single H200 GPU.}\vspace{-2mm}
\label{tab:pixelflow}
\centering
\resizebox{0.9\linewidth}{!}{

\begin{tabular}{l|c|c|c}
\toprule
Attention & FID & Inception Score & Throughput \\
\midrule 
VSA & 23.59 & 64.07 & 32.30 \\
SLA & 22.58 & 65.31 & 29.81 \\
LLSA & \textbf{20.41} & \textbf{73.21} & \textbf{34.16} \\
\bottomrule
\end{tabular}}\vspace{-0mm}
\end{table}

\noindent \textbf{ImageNet-256 Benchmark.}
We further evaluate LLSA on the ImageNet \(256\times256\) dataset using PixelFlow~\cite{chen2025pixelflow}, a multi-stage pixel diffusion model that progressively upsamples a noisy image from a low-resolution token sequence to higher resolutions. On ImageNet-256, it employs a $4 \times 4$ patchification and upsamples an $ 8 \times 8$ token image into $ 64 \times 64$ images via a four-stage diffusion process. We replace the full attention with sparse attention methods only at the highest-resolution stage, as lower-resolution stages (\(\le 32\times 32\)) do not benefit from sparsification. All methods are trained for $10$ epochs to ensure a fair comparison.

As shown in Table~\ref{tab:pixelflow}, LLSA again outperforms VSA and SLA in both FID and Inception Score, reinforcing its advantage on a more challenging dataset and demonstrating its potential to accelerate large-scale diffusion models.

Overall, our evaluation setup is conservative: we allocate baselines a larger effective $K$, ensuring that LLSA improves performance not due to experimental bias but due to its more effective and computationally efficient attention formulation. More qualitative results are given in the supplementary material.

\subsection{Efficiency Benchmark}

\begin{figure*}[t]
\centering
\includegraphics[width=0.8\linewidth]{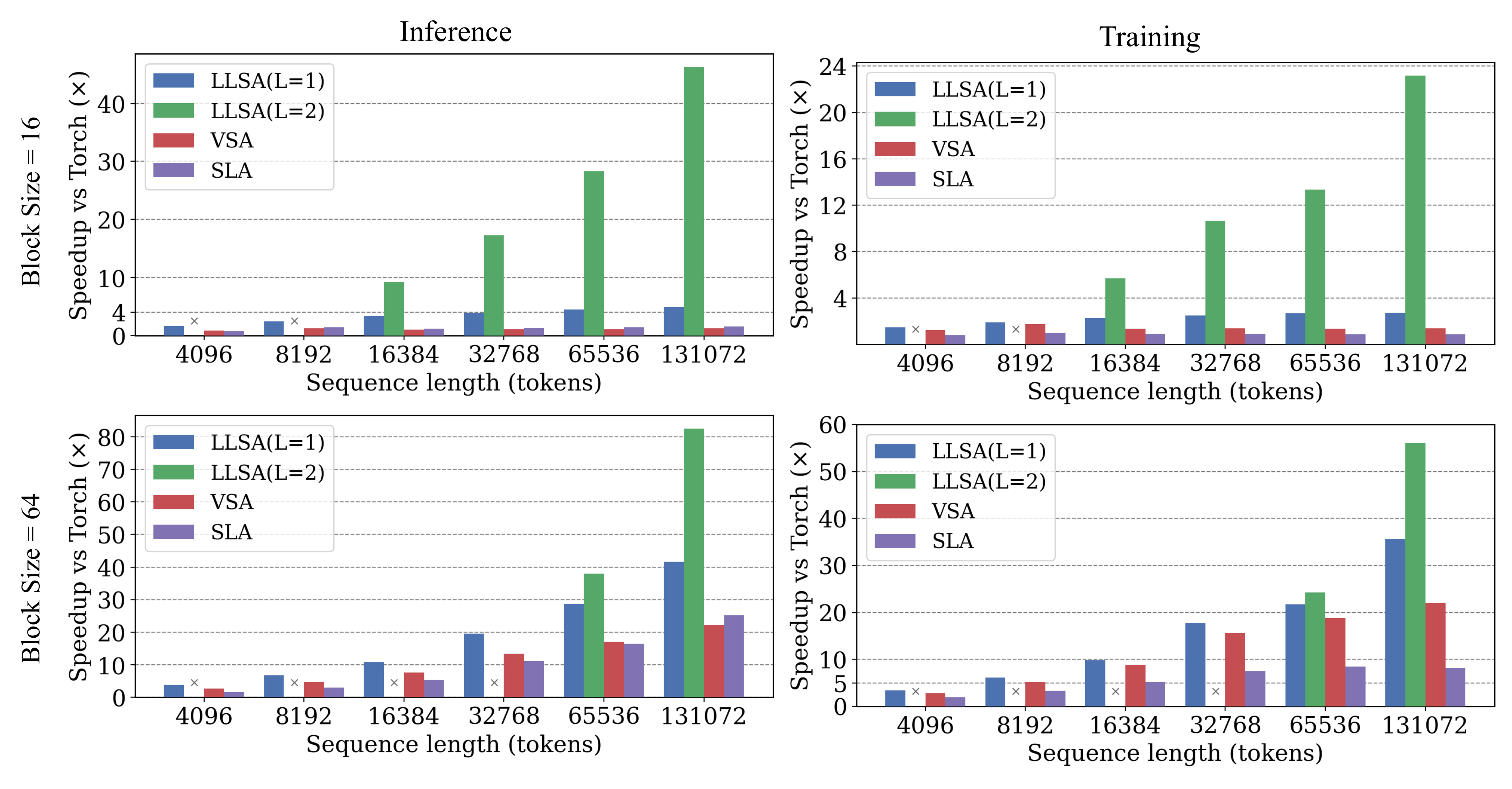}\vspace{-3mm}
\caption{Acceleration ratio of different attention methods compared to PyTorch Attention (FlashAttention2). We evaluate training and inference with block size $B \in \{16,64\}$ across varying sequence lengths on an H200 GPU.}\vspace{-4mm}
\label{fig:attention_ratio}
\end{figure*}

\noindent \textbf{Attention Performance}
Figure~\ref{fig:attention_ratio} compares the inference and training speed of LLSA with VSA and SLA.  
LLSA is implemented in Triton~\cite{tillet2019triton}, and VSA/SLA use their official Triton kernels.  
For a fair comparison, we set $K=8$ for LLSA and increase $K$ for VSA and SLA such that all methods process the same number of KV blocks during sparse attention (Sec.~\ref{sec:comparison}).

Across all settings, LLSA consistently outperforms VSA and SLA, often by a substantial margin.  
The efficiency gains arise from three complementary factors: (1) LLSA uses a more efficient sparse indices transpose algorithm rather than a mask-based algorithm to compute the gradient of key and value. (2) Due to Hierarchical KV Enrichment, LLSA retrieves the same number of relevant tokens with a smaller $K$, reducing the overhead of Top-$K$ selection.  (3) When the sequence becomes sufficiently long and $L=1$ is constrained by quadratic cost, switching to $L=2$ yields a clear improvement, reflecting the expected log-linear scaling.

For a block size of $B=16$, VSA and SLA exhibit significant kernel-level inefficiencies due to the quadratic runtime on both Top-$K$ selection and mask-based backward algorithm.  
With $B=64$, their kernels become more competitive, yet LLSA still maintains a clear advantage, especially at long sequence lengths where the selection stage becomes dominant.

\begin{figure}[t]
\centering
\includegraphics[width=0.8\linewidth]{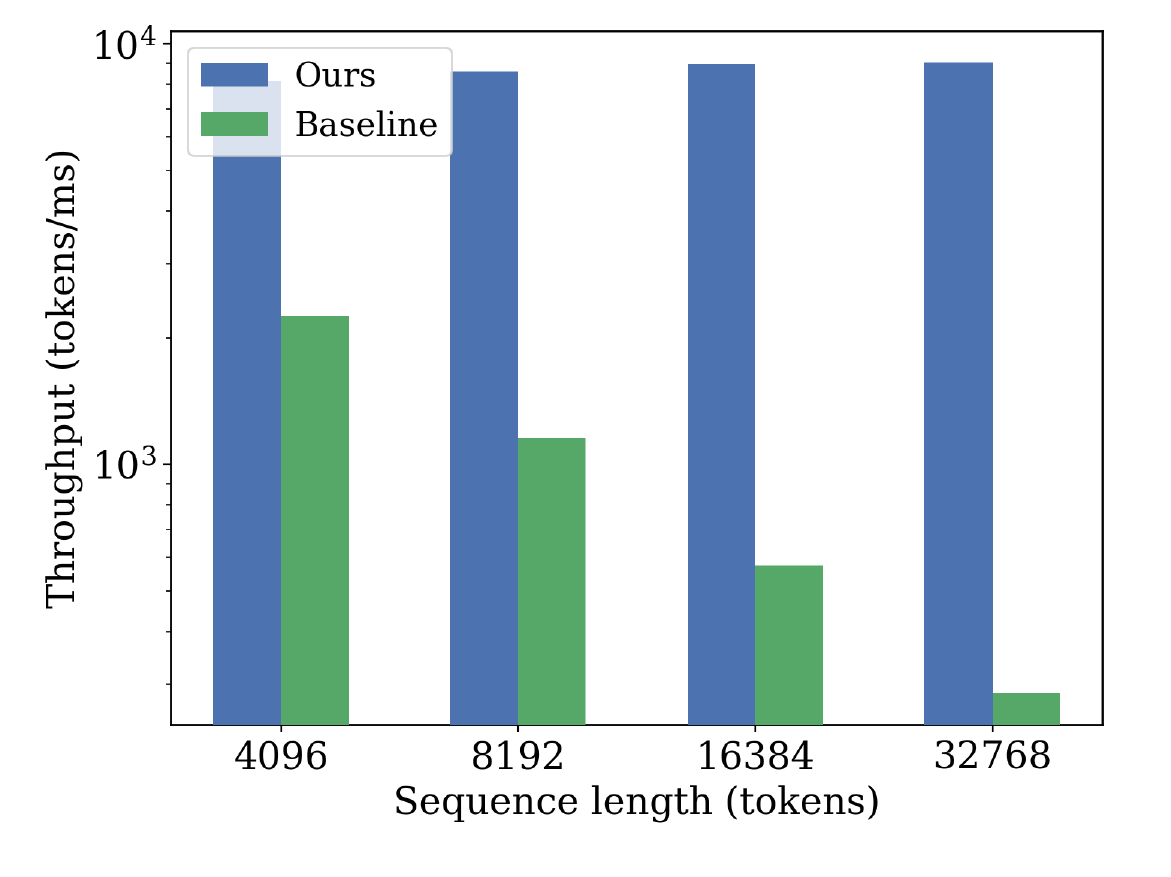}\vspace{-3mm}
\caption{The throughput of sparse key-value backward. Experiments are conducted on an H200 GPU using tokens with $64$ heads and head dimension $64$. We set $K=8$ and $B=16$ for sparse Top-$K$ attention.}
\vspace{-0mm}
\label{fig:backward}
\end{figure}

\noindent \textbf{Efficient Backward Implementation}
We further compare our key–value backward kernel, which operates directly on sparse CSC indices obtained by the sparse-indices transpose operator, with a mask-based baseline widely used in prior work~\cite{sla}.  
Both methods use the same Top-$K$ indices obtained from a standard forward pass.
Our implementation includes a sparse indices transpose kernel and a key-value backward kernel with CSC indices. For the baseline implementation, we convert the sparse indices into a dense binary mask and run a standard block-sparse backward algorithm that skips the unmarked blocks.

As shown in Fig.~\ref{fig:backward}, our implementation maintains nearly constant throughput across all sequence lengths, confirming its \textbf{linear complexity}.  
In contrast, the baseline must convert sparse indices into a dense block mask, an $O(N^2)$ operation, and its block-sparse backward must scan this dense mask to skip unselected blocks. This results in a steady degradation in throughput as the sequence length increases.

Overall, the results demonstrate that LLSA achieves superior efficiency in attention computation, further validating its suitability for scaling transformers to extremely long sequences.
\section{Conclusion}
\label{sec:conclusion}
We present Log-linear Sparse Attention, a novel attention mechanism that reduces the computational complexity of self-attention from $O(N^2)$ to $O(N \log N)$. LLSA extends the previous single-level search strategy to a hierarchical search procedure and incorporates Hierarchical KV Enrichment to preserve global context and maintain model quality under long-sequence sparsification. The hierarchical design allows us to outperform previous sparse attention methods with a smaller $K$. In addition, we develop a high-performance GPU implementation based on block sparse attention and an efficient Top-$K$ indices backward algorithm that eliminates the need for dense masks. Experiments on high-resolution pixel-space DiTs demonstrate that LLSA achieves a better FID score and higher training throughput than prior sparse attention methods, indicating its potential for scaling diffusion transformers to substantially longer token sequences.

\small{
\noindent
\textbf{Acknowledgments.} This research is supported by NTU SUG-NAP. This research is also supported by cash and in-kind funding from NTU S-Lab and industry partner(s).}

{
    \small
    \bibliographystyle{ieeenat_fullname}
    \bibliography{main}
}

\setcounter{section}{0}
\renewcommand\thesection{\Alph{section}}

\section{Implementation Details}

\subsection{Clarification of Kernel Implementation}

Both SLA~\cite{sla} and VSA~\cite{vsa} use inefficient sparse backward implementations. SLA applies the standard mask-based sparse block sparse attention backward, skipping the unused queries for each key. VSA implements a preprocessing kernel that extracts query indices for each key from the binary sparse mask, but it still requires $O(N^2)$ complexity to construct the binary mask. 

In Table~\ref{tab:comparison}, to fairly evaluate the algorithms' throughput rather than their GPU implementations, we reimplement their sparse backward kernels with our efficient sparse index transpose kernel. This conservative experimental setting significantly improves the baselines' throughput, but results indicate that LLSA is still more efficient than SLA and VSA for an equal number of effective sparse tokens.

In Fig.~\ref{fig:attention_ratio}, we compare the kernel efficiency of LLSA with the official implementations of SLA and VSA. We modify their block size configurations to support $B=16$ inference.  

\subsection{Training Configuration}

\begin{table} []
\caption{Hyperparameters of Pixel DiT trained on FFHQ and ImageNet of various resolutions. Models with different attention implementations have identical configurations. FFHQ models are trained on one H200 GPU and ImagetNet models are trained on four H200 GPUs.}\vspace{-2mm}
\label{tab:hyperparameters}
\resizebox{\linewidth}{!}{
\centering
\begin{tabular}{l|c|c|c|c|c|c}
\toprule
Model & FFHQ-32  & FFHQ-128  & FFHQ-256 & FFHQ-512 & ImageNet-128 & ImageNet-256 \\
\midrule 
Patch Size & $1 \times 1$ & $1 \times 1$ & $1 \times 1$ & $1 \times 1$ & $4\times4$ & $4\times4$  \\
DiT Config & DiT-S & DiT-S & DiT-S & DiT-S & PixelFlow-L & PixelFlow-L \\
Pretrained Model & - & FFHQ-32 & FFHQ-128 & FFHQ-256 & - & ImageNet-128\\
SNR Rescale & 1 & 2 & 4 & 8 & 1 & 1\\
Epochs & 40 & 20 & 10 & 2 & 40 & 10\\
Batch Size & 64 & 16 & 4 & 1 & 32 & 8 \\
Learning Rate & $1 \times 10^{-4}$ & $1 \times 10^{-4}$ & $1 \times 10^{-4}$ & $1 \times 10^{-4}$ & $1 \times 10^{-4}$ & $1 \times 10^{-4}$ \\
\bottomrule
\end{tabular}}\vspace{-2mm}
\end{table}

We provide the detailed training configurations on FFHQ~\cite{stylegan} and ImageNet~\cite{imagenet} in Table~\ref{tab:hyperparameters}.

\section{Additional Experiment Results}

\subsection{Additional Ablation Results}

Similar to the ablation studies presented in Sec.~\ref{sec:abl}, all experiments in this section are conducted using DiT-S trained on $128\times128$ FFHQ unless otherwise specified. Table~\ref{tab:abl_6} and Table~\ref{tab:abl_7} report results after 10 epochs of training.

\begin{table} [t]
\caption{Ablation study results of Log-linear Sparse Attention}\vspace{-2mm}
\label{tab:abl_supp}

\centering
\subfloat[Enrichment Levels]{%
  \begin{minipage}[t]{0.8\linewidth}
\centering
\resizebox{\linewidth}{!}{
\begin{tabular}{l|c|c|c}
\toprule
Configuration & $L_e$ & FID & Throughput  \\
\midrule 
LLSA FFHQ-128 ($L=2$) & 0 & 27.98 & 500.38 \\
LLSA FFHQ-128 ($L=2$) & 1 & 25.49 & 467.35 \\
LLSA FFHQ-128 ($L=2$) & 2 & 24.37 & 436.40 \\

\bottomrule
\end{tabular}
}
\end{minipage}
\label{tab:abl_4}
}
\par\vspace{0.5em}
\centering
\subfloat[Extension to $512\times512$ Resolution]{%
  \begin{minipage}[t]{0.8\linewidth}
\centering
\resizebox{\linewidth}{!}{
\begin{tabular}{l|c|c}
\toprule
Configuration & FID & Throughput  \\
\midrule 
LLSA FFHQ-256 ($L=2$) & 39.29 & 375.34 \\
\midrule 
LLSA FFHQ-512 ($L=1$) & - & 44.90 \\
LLSA FFHQ-512 ($L=2$) & 39.26 & 292.66 \\
LLSA FFHQ-512 ($L=3$) & 40.77 & 323.29 \\

\bottomrule
\end{tabular}
}
\end{minipage}
\label{tab:abl_5}
}
\par\vspace{0.5em}
\hspace{1.5 em}
\subfloat[SNR Adjusting Methods]{
  \begin{minipage}[t]{0.4\linewidth}
\centering
\resizebox{\linewidth}{!}{
\begin{tabular}{l|c}
\toprule
Configuration & FID  \\
\midrule 
Baseline & 32.66  \\
Timestep Shift & 30.32  \\
Logit-normal & 30.22  \\
Noise Rescale & \textbf{29.46} \\
\bottomrule
\end{tabular}
}
\end{minipage}
\label{tab:abl_6}
}
\hfill
\centering
\subfloat[Index Reordering]{
  \begin{minipage}[t]{0.4\linewidth}
\centering
\resizebox{\linewidth}{!}{
\begin{tabular}{l|c}
\toprule
Configuration & FID  \\
\midrule 
Raster Order & 31.19  \\
Index Reordering & \textbf{29.46}  \\
\bottomrule
\end{tabular}
}
\end{minipage}
\label{tab:abl_7}
}
\hspace{1.0em}
\end{table}

\noindent \textbf{More enrichment levels lead to better quality.}
We train three two-level LLSA DiT on FFHQ-128 with different KV enrichment layers $L_e$, shown in Table~\ref{tab:abl_4}. More enrichment levels increase the effective token number and generation quality, while slightly reducing throughput.

\noindent \textbf{$512\times512$ pixel token sequence generation.}
To assess the scalability of LLSA on substantially longer token sequences, we train DiT-S on FFHQ-512 using different numbers of hierarchical levels $L$ (Table~\ref{tab:abl_5}). The single-level LLSA ($L=1$) fails to converge within a reasonable time budget due to its $O(N^2)$ selection cost and coarse tokens. Increasing to $L=2$ dramatically improves throughput. Further extending to $L=3$ yields additional speed gains. The scaling of per-token throughput, from $375.34$ (at $256\times256$) to $323.29$ (at $512\times512$), closely follows the $O(N \log N)$ complexity of LLSA. 

\noindent \textbf{Noise rescaling is the most effective SNR-adjustment method.}
We evaluate three approaches for adjusting the SNR when training on higher-resolution images.  
(1) \emph{Timestep Shift:} Following~\cite{sd3}, we apply a timestep shift to align the SNR of images with sequence length $m$ to those of length $n$: 
$t_m = \frac{\sqrt{m/n}\, t_n}{1 + (\sqrt{m/n}-1)t_n}$.  
Since our target SNR corresponds to $64\times64$ images, we set $\sqrt{m/n}=2$ for the $128\times128$ experiments.  
(2) \emph{Logit-Normal Sampler:} As proposed in~\cite{sd3}, we replace uniform timestep sampling with a logit-normal sampler using its default parameters ($\mu=1$, $s=1.0$).  
(3) \emph{Noise Rescaling:} As described in Sec.~\ref{sec:exp_imple}, we rescale the noise by a factor of $s=\sqrt{m/64^2}$ for images of resolution $m$, thereby matching their effective SNR to that of $64\times64$ images.  
Results in Table~\ref{tab:abl_6} show that noise rescaling yields the best generation quality among the tested methods.

\noindent \textbf{Index reordering improves model quality.}
To assess the impact of index reordering, we train an additional model using the default raster-scan ordering for 2D pixels.  
As shown in Table~\ref{tab:abl_7}, the model without index reordering can still converge but achieves worse FID, confirming the usefulness of the spatially coherent index reordering.

\begin{figure}[t]
\centering
\includegraphics[width=0.9\linewidth]{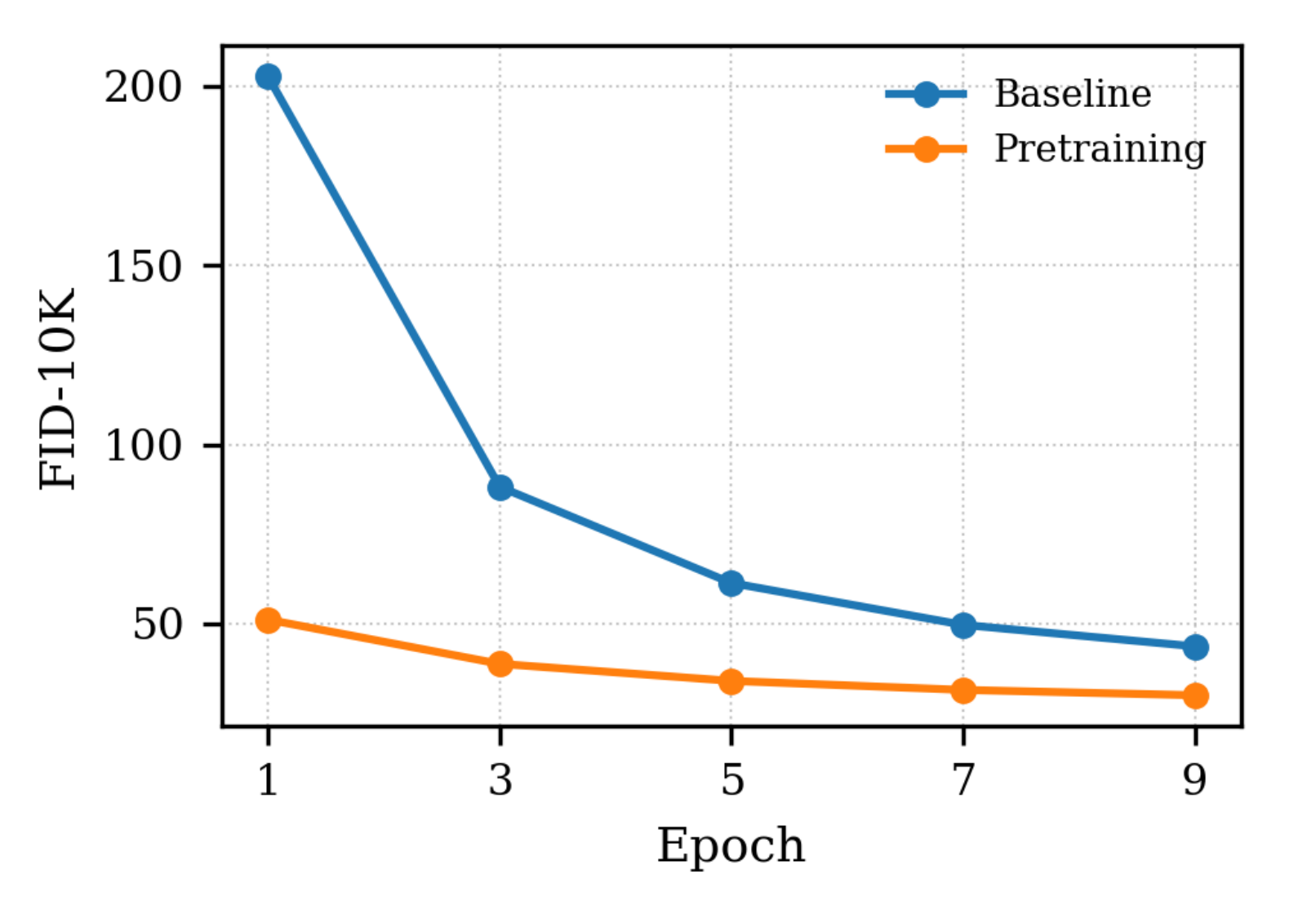}\vspace{-3mm}
\caption{The FID curves of different training strategies. Compared to training from scratch, starting from a model pretrained on low-resolution data significantly reduces training cost.}
\vspace{-0mm}
\label{fig:pretraining}
\end{figure}

\noindent \textbf{Pretraining substantially reduces training cost.}
Figure~\ref{fig:pretraining} compares the FID curves of a model trained from scratch and one initialized from a lower-resolution pretrained checkpoint.  
The pretrained model converges rapidly—within the first epoch—demonstrating that low-resolution pretraining significantly accelerates high-resolution pixel-space DiT training.

\subsection{Training Curves}

\begin{figure}[t]
\centering
\includegraphics[width=\linewidth]{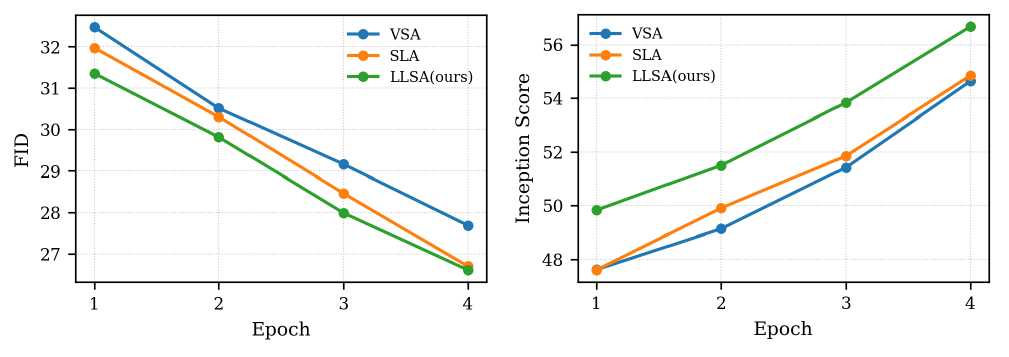}\vspace{-3mm}
\caption{The FID and Inception Score curves of the first $4$ epochs using VSA, SLA, and LLSA on PixelFlow ImageNet-256 benchmark.}
\vspace{-0mm}
\label{fig:curve}
\end{figure}

We show the training FID~\cite{fid} and Inception Score~\cite{is} curves on PixelFlow~\cite{chen2025pixelflow} ImageNet-256 benchmark in Fig.~\ref{fig:curve}. The generation quality of LLSA is consistently better than that of baseline attention approaches throughout the training process. 

\section{Qualitative Results}

\begin{figure}[t]
\centering
\includegraphics[width=\linewidth]{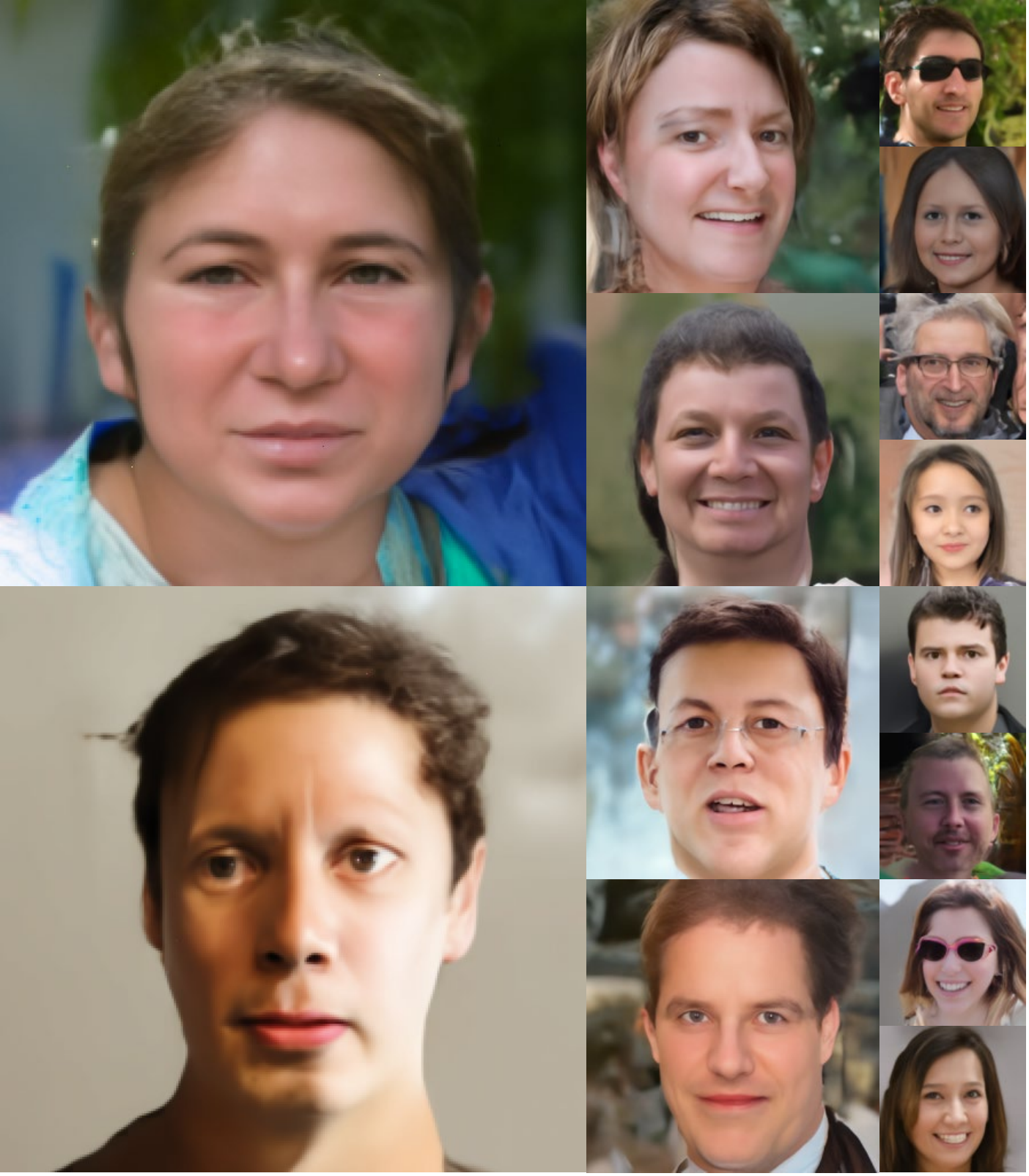}\vspace{-3mm}
\caption{The qualitative results of pixel space DiT-S using LLSA trained on FFHQ-128, FFHQ-256, and FFHQ-512. For FFHQ-512, the model is only trained for two epochs. We believe that better quality can be obtained by longer training.}
\vspace{-0mm}
\label{fig:ffhq_img}
\end{figure}

\begin{figure}[t]
\centering
\includegraphics[width=\linewidth]{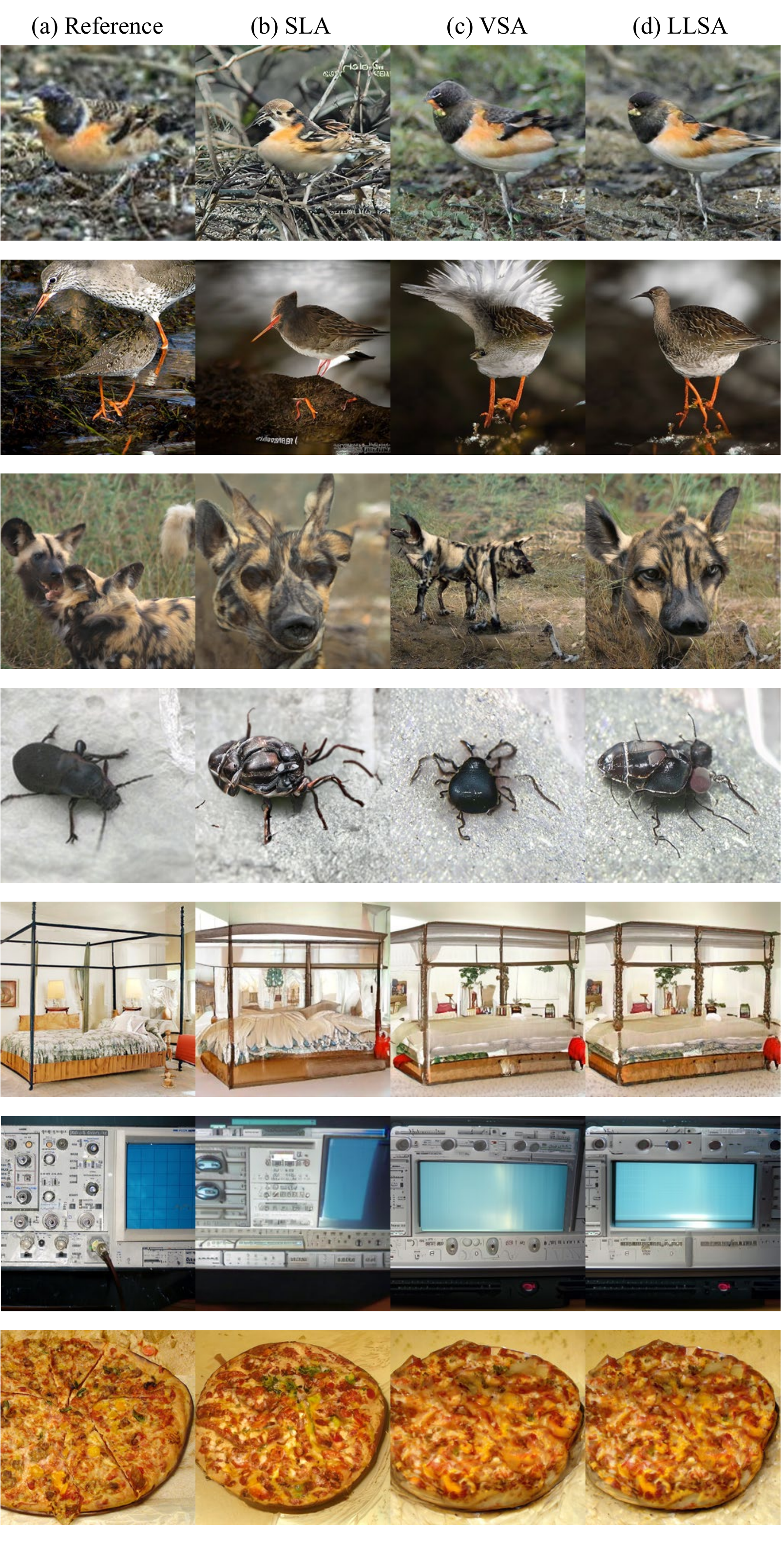}\vspace{-3mm}
\caption{The qualitative comparison of SLA, VSA, and LLSA trained on PixelFlow-L ImageNet-256. The reference images are generated by a well-trained full-attention PixelFlow model from the official repository.}
\vspace{-0mm}
\label{fig:imagenet_img}
\end{figure}

We present qualitative results of LLSA in this section. In Fig.~\ref{fig:ffhq_img}, we show the samples generated from LLSA DiT-S trained on FFHQ-128, FFHQ-256, and FFHQ-512. For FFHQ-512, the model is only trained for 2 epochs. We believe that better quality can be obtained by longer training. In Fig.~\ref{fig:imagenet_img}, we compare the ImageNet-256 samples generated by LLSA PixelFlow-L with the those produced by SLA and VSA variants. For reference, we also include samples generated by the official best-performing PixelFlow model.
\newpage

\clearpage

\end{document}